\documentclass[journal]{IEEEtran}
\ifCLASSINFOpdf
\else
\fi
\usepackage{hyperref}

\usepackage{tabularx}
\usepackage{times}
\usepackage{epsfig}
\usepackage{graphicx}
\usepackage{amsmath}
\usepackage{amssymb}
\usepackage{subfig}
\usepackage{multirow}
\usepackage{url}
\usepackage{color,soul}
\usepackage{bbold}
\usepackage[linesnumbered,ruled,vlined]{algorithm2e} 

\usepackage{setspace}

\usepackage[numbers,sort]{natbib}
\usepackage[colorinlistoftodos]{todonotes}

\begin{document}
%
\title{Player Tracking and Identification in Ice Hockey}
%
%
%

\author{Kanav Vats, ~\IEEEmembership{Systems Design Engineering, University of Waterloo}\\ 
        Pascale Walters, ~\IEEEmembership{Stathletes Inc.}\\
        Mehrnaz Fani, ~\IEEEmembership{Systems Design Engineering, University of Waterloo}\\
       
        David A. Clausi, ~\IEEEmembership{Systems Design Engineering, University of Waterloo}\\
        John S. Zelek ~\IEEEmembership{Systems Design Engineering, University of Waterloo}

}

%
%

\markboth{Journal of \LaTeX\ Class Files,~Vol.~14, No.~8, August~2015}%
{Shell \MakeLowercase{\textit{et al.}}: Bare Demo of IEEEtran.cls for IEEE Journals}
%



\maketitle

\begin{abstract}
Tracking and identifying players is a fundamental step in computer vision-based ice hockey analytics. The data generated by tracking is used in many other downstream tasks, such as game event detection and game strategy analysis. Player tracking and identification is a challenging problem since the motion of players in hockey is fast-paced and non-linear when compared to pedestrians. There is also significant camera panning and zooming in hockey broadcast video. Identifying players in ice hockey is challenging since the players of the same team appear almost identical, with the jersey number the only consistent discriminating factor between players. To address this problem, an automated system to track and identify players in broadcast NHL hockey videos is introduced. The system is composed of three components (1) player tracking, (2) team identification and (3) player identification. Due to the absence of publicly available datasets, the datasets used to train the three components are annotated manually. Player tracking is performed with the help of a state of the art tracking algorithm obtaining a Multi-Object Tracking Accuracy (MOTA) score of $94.5\%$. For team identification, away-team jerseys are grouped into a single class and home-team jerseys are grouped in classes according to their jersey color. A convolutional neural network is then trained on the team identification dataset. The team identification network obtains an accuracy of $97\%$ on the test set.  A novel player identification model is introduced that utilizes a temporal one-dimensional convolutional network to identify players from player bounding box sequences. The player identification model further takes advantage of the available NHL game roster data to obtain a player identification accuracy of $83\%$.
\end{abstract}


\section{Introduction}

Ice hockey is a popular sport played by millions of people \cite{iihf}. Being a team sport, knowing the location of players on the ice rink is essential for analyzing the game strategy and player performance. The locations of the players on the rink during the game are used by coaches, scouts, and statisticians for analyzing the play. Although player location data can be obtained manually, the process of labelling data by hand on a per-game basis is extremely tedious and time consuming. Therefore, an automated computer vision-based player tracking and identification system is of high utility.

In this paper, we introduce an automated system to track and identify players in broadcast National Hockey League (NHL) videos. Referees, being a part of the game, are also tracked and identified separately from players. The input to the system is broadcast NHL clips from the main camera view (i.e., camera located in the stands above the centre ice line) and the output are player trajectories along with their identities. Since there are no publicly available datasets for ice hockey player tracking, team identification, and player identification, we annotate our own datasets for each of these problems. The previous papers in ice hockey player tracking \cite{cai_hockey, okuma} make use of hand crafted features for detection and re-identification. Therefore, we perform experiments with five state of the art tracking algorithms  \cite{Braso_2020_CVPR,tracktor_2019_ICCV,Wojke2017simple,Bewley2016_sort, zhang2020fair} on our hockey player tracking dataset and evaluate their performance. The output of the player tracking algorithm is a temporal sequence of player bounding boxes, called player tracklets.

\begin{figure}
    \centering
    \includegraphics[scale=0.55]{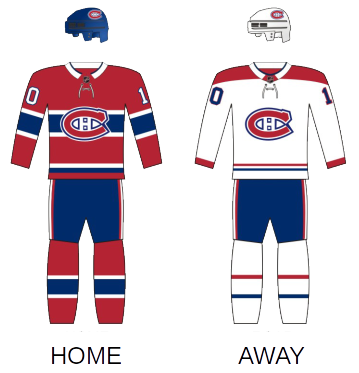}
    \caption{Home (dark) and away (white) jerseys worn by the Montreal Canadiens of the National Hockey League \cite{martello_2020}.}
    \label{fig:mtl_jerseys}
\end{figure}

Posing team identification as a classification problem with each team treated as a separate class would be impractical since (1) this will result in a large number of classes, and (2) the same NHL team wears two different colors based on whether it is the home or away team (Fig. \ref{fig:mtl_jerseys}). Therefore, instead of treating each team as a separate class, we treat the away (light) jerseys of all teams as a single class and cluster home jerseys based on their jersey color. Since referees are easily distinguishable from players, they are treated as a separate class.Based on this simple training data formation, hockey players can be classified into home and away teams. The team identification network obtains an accuracy of $96.6\%$ on the test set and does not require additional fine tuning on new games.


\begin{figure*}[!th]
\begin{center}
\includegraphics[width=\linewidth]{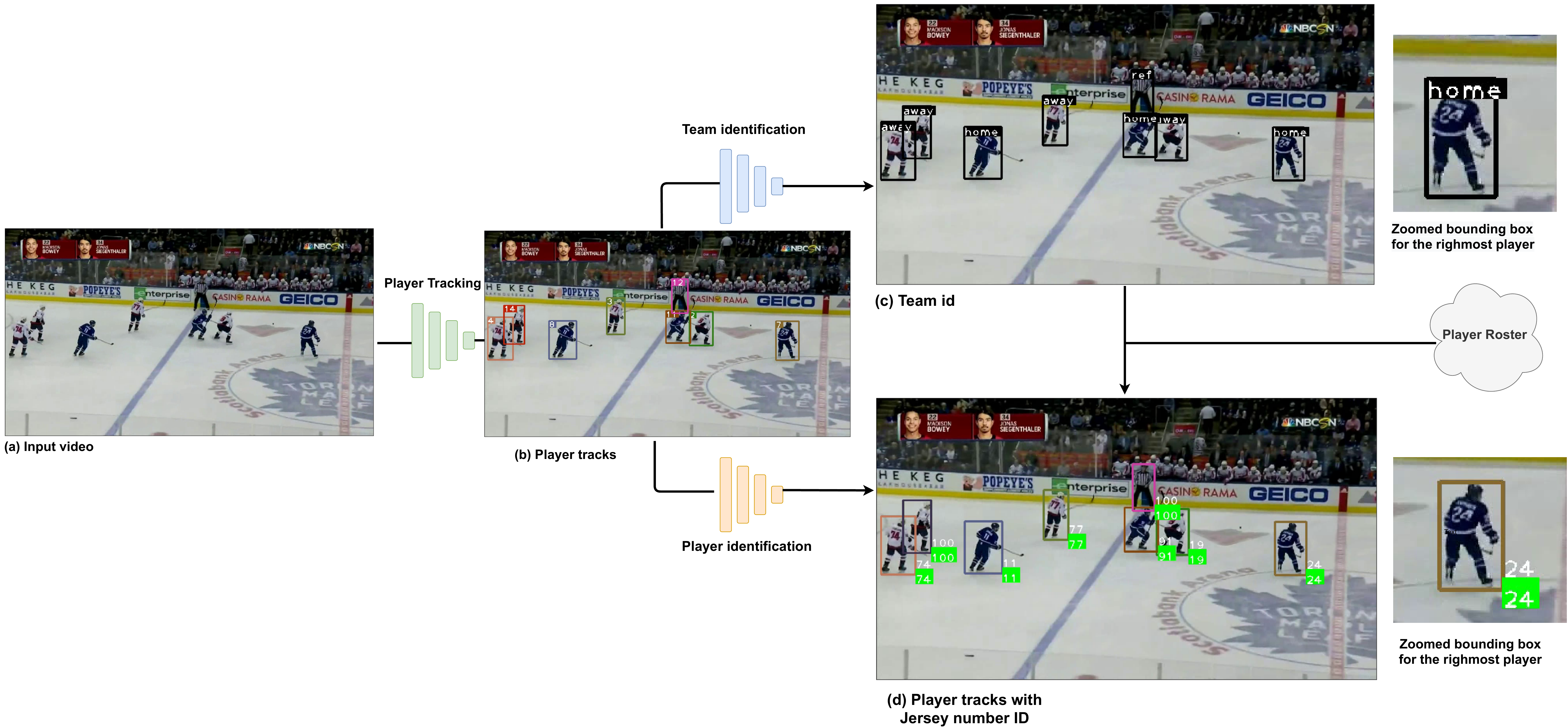}
\end{center}
  \caption{ Overview of the player tracking and identification system. The tracking model takes a hockey broadcast video clip as input and outputs player tracks. The team identification model takes the player track bounding boxes as input and identifies the team of each player along with identifying the referees. The player identification model utilizes the player tracks, team data and game roster data to output player tracks with jersey number identities. }
\label{fig:pipeline}
\end{figure*}

Unlike soccer and basketball \cite{Senocak_2018_CVPR_Workshops} where player facial features and skin color are visible, a big challenge in player identification in hockey is that the players of the same team appear nearly identical due to having the same uniform as well as having similar physical size.. Therefore, we use jersey number for identifying players since it is the most prominent feature present on all player jerseys. Instead of classifying jersey numbers from static images \cite{gerke,li,liu}, we identify a player's jersey number from player tracklets. Player tracklets allow a model to process temporal context to identify a jersey number since that is likely to be visible in multiple frames of the tracklet.  We introduce a temporal 1-dimensional convolutional neural network (1D CNN)-based network for identifying players from their tracklets. The network generates a higher accuracy than the previous work by Chan \textit{et al.} \cite{CHAN2021113891} by $9.9\%$ without requiring any additional probability score aggregation model for inference.

The tracking, team identification, and player identification models are combined to form a holistic offline system to track and identify players and referees in the broadcast videos. Player tracking helps team identification by removing team identification errors in player tracklets through a simple majority voting. Additionally, based on the team identification output, we use the game roster data to further improve the identification performance of the automated system by an additional $5\%$. The overall system is depicted in Fig. \ref{fig:pipeline}. The system is able to identify players from video with an accuracy of $82.8\%$ with a Multi-Object Tracking Accuracy (MOTA) score of $94.5\%$ and an Identification $F_1$ (IDF1) score of $62.9\%$.

Five computer vision contributions are recognized applied to the game of ice hockey: 
\begin{enumerate}
    \item New ice hockey datasets are introduced for player tracking, team identification, and player identification from tracklets.
    \item We compare and contrast several state-of-the-art tracking algorithms and analyze their performance and failure modes.
    \item A simple but efficient team identification algorithm for ice hockey is implemented.
    \item A temporal 1D CNN based player identification model is implemented that outperforms the current state of the art \cite{CHAN2021113891} by $9.9\%$.
    \item A holistic system that combines tracking, team identification, and player identification models, along with making use of the team roster data, to track and identify players in broadcast ice hockey videos is established.
\end{enumerate}

\section{Background}
\begin{figure*}[t]
\begin{center}
\includegraphics[width=\linewidth]{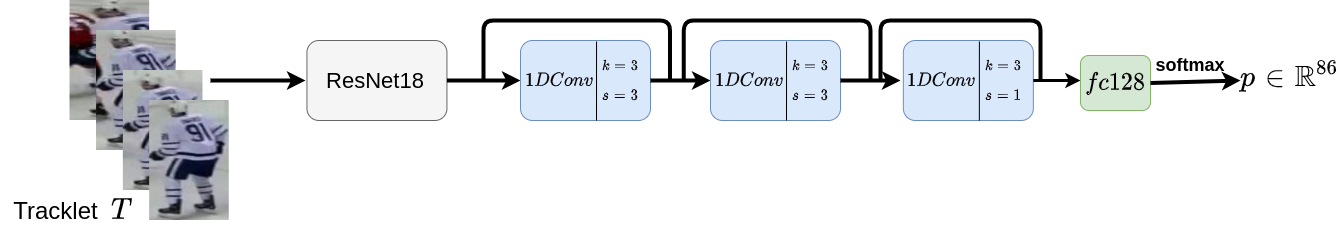}
\end{center}
  \caption{ Network architecture for the player identification model. The networks accepts a player tracklet as input. Each tracklet image is passed through a ResNet18 to obtain time ordered features $F$. The features $F$ are input into three 1D convolutional blocks, each consisting of a 1D convolutional layer, batch normalization, and ReLU activation. In this figure, $k$ and $s$ are the kernel size and stride of convolution operation. The activations obtained from the convolutions blocks are mean-pooled and passed through a fully connected layer and a softmax layer to output the probability distribution of jersey number $p \in \mathbb{R}^86$. }
\label{fig:jn_arch}
\end{figure*}

\subsection{Tracking}
The objective of multi-object tracking (MOT) is to detect objects of interest in video frames and associate the detections with appropriate trajectories. Player tracking is an important problem in computer vision-based sports analytics, since player tracking combined with an automatic homography estimation system \cite{jiang2020optimizing} is used to obtain absolute player locations on the sports rink. Also, various computer vision-based tasks, such as sports event detection \cite{Vats_2020_CVPR_Workshops,Sanford_2020_CVPR_Workshops,Vats_2021_CVPR}, can be improved with player tracking data.

Tracking by detection (TBD) is a widely used approach for multi-object tracking. Tracking by detection consists of two steps: (1) detecting objects of interest (hockey players in our case) frame-by-frame in the video, then (2) linking player detections to produce tracks using a tracking algorithm. Detection is usually done with the help of a deep detector, such as Faster R-CNN \cite{NIPS2015_14bfa6bb} or YOLO \cite{yolo}. For associating detections with trajectories, techniques such as Kalman filtering with Hungarian algorithm \cite{Bewley2016_sort, Wojke2017simple, zhang2020fair} and graphical inference \cite{Braso_2020_CVPR,tanglifted} are used. In recent literature, re-identification in tracking is commonly carried out with the help of deep CNNs using appearance \cite{Wojke2017simple, Braso_2020_CVPR,zhang2020fair} and pose features  \cite{tanglifted}.

For sports player tracking, Sanguesa \textit{et al.} \cite{Sangesa2019SingleCameraBT} demonstrated that deep features perform better than classical hand crafted features for basketball player tracking. Lu \textit{et al.} \cite{lu} perform player tracking in basketball using a Kalman filter. Theagarajan \textit{et al.} \cite{Theagarajan_2021} track players in soccer videos using the DeepSORT algorithm \cite{Wojke2017simple}. Hurault \textit{et al} \cite{hurault} introduce a self-supervised detection algorithm to detect small soccer players and track players in non-broadcast settings using a triplet loss trained re-identification mechanism, with embeddings obtained from the detector itself.

 In ice hockey, Okuma \textit{et al.} \cite{okuma} track hockey players by introducing a particle filter combined with mixture particle filter (MPF) framework \cite{mixture_tracking}, along with an Adaboost \cite{adaboost} player detector. The MPF framework \cite{mixture_tracking} allows the particle filter framework to handle multi-modality by modelling the posterior state distributions of $M$ objects as an $M$ component mixture. A disadvantage of the MPF framework is that the particles merge and split in the process and leads to loss of identities. Moreover, the algorithm did not have any mechanism to prevent identity switches and lost identities of players after occlusions. Cai \textit{et al.} \cite{cai_hockey} improved upon \cite{okuma} by using a bipartite matching for associating observations with targets instead of using the mixture particle framework. However, the algorithm is not trained or tested on broadcast videos, but performs tracking in the rink coordinate system after a manual homography calculation.
 
 In ice hockey, prior published research \cite{okuma,cai_hockey} perform player tracking with the help of handcrafted features for player detection and re-identification. In this paper we track and identify hockey players in broadcast NHL videos and analyze performance of several state-of-the-art deep tracking models on the ice hockey dataset.

\subsection{Player Identification}
Identifying players and referees is one of the most important problems in computer vision-based sports analytics. Analyzing individual player actions and player performance from broadcast video is not feasible without detecting and identifying the player. Before the advent of deep learning methods, player identification was performed with the help of handcrafted features \cite{Saric2008PlayerNL}.  Although techniques for identifying players from body appearance exist \cite{Senocak_2018_CVPR_Workshops}, jersey number is the primary and most widely used feature for player identification, since it is observable and consistent throughout a game. Most deep learning based player identification approaches in the literature focus on identifying the player jersey number from single frames using a CNN \cite{gerke, li, liu}. Gerke \textit{et al.} \cite{gerke} were one of the first to use CNNs for soccer jersey number identification and found that deep learning approach outperforms handcrafted features. Li \textit{et al.} \cite{li} employed a semi-supervised spatial transformer network to help the CNN localize the jersey number in the player image. Liu \textit{et al.} \cite{liu} use a pose-guided R-CNN for jersey digit localization and classification by introducing a human keypoint prediction branch to the network and a pose-guided regressor to generate digit proposals. Gerke \textit{et al.} \cite{GERKE2017105} also combined their single-frame based jersey classifier with soccer field constellation features to identify players. Vats \textit{et al.} \cite{Vats2021MultitaskLF} use a multi-task learning loss based approach to identify jersey numbers from static images.

Zhang \textit{et al.} \cite{ZHANG2020107260} track and identify players in a multi-camera setting using a distinguishable deep representation of player identity using a coarse-to-fine framework. Lu \textit{et al.} \cite{lu} use a variant of Expectation-Maximization (EM) algorithm to learn a Conditional Random Field (CRF) model composed of player identity and feature nodes. Chan \textit{et al.} \cite{CHAN2021113891} use a combination of a CNN and long short term memory network (LSTM) \cite{lstm} similar to the long term recurrent convolutional network (LRCN) by Dohnaue \textit{et al.} \cite{lrcn2014} for identifying players from player sequences. The final inference in Chan \textit{el al.} \cite{CHAN2021113891} is performed using a another CNN network applied over probability scores obtained from CNN LSTM network.

In this paper, we identify players using player tracklets with the help of a temporal 1D CNN. Our proposed inference scheme does not require the use of an additional network.

\subsection{Team Identification}

Beyond knowing the identity of a player, the player must also be assigned to a team. Many sports analytics, such as ``shot attempts" and ``team formations", require knowing the team to which each individual belongs. In sports leagues, teams differentiate themselves based on the colour and design of the jerseys worn by the players. In ice hockey, formulating team identification as a classification problem with each team treated as a separate class is not feasible since teams use different colored jerseys depending if they are the 'home' or 'away' team. Teams wear light- and dark-coloured jerseys depending on whether they are playing at their home venue or away venue (Fig. \ref{fig:mtl_jerseys}).  Furthermore, each game in which new teams play would require fine-tuning \cite{Koshkina_2021_CVPR}.


Early work used colour histograms or colour features with a clustering approach to differentiate teams \cite{shitrit2011, ivankovic2014,dorazio2009, lu2013, liu2014, bialkowski2014, tong2011, guo2020, mazzeo2008, ajmeri2018}. This approach, while being lightweight, does not address occlusions, changes in illumination, and teams wearing similar jersey colours  \cite{shitrit2011, Koshkina_2021_CVPR}. Deep learning approaches have increased performance and generalizablitity of player classification models \cite{Istasse_2019_CVPR_Workshops}.

Istasse \textit{et al.} \cite{Istasse_2019_CVPR_Workshops} simultaneously segment and classify players in indoor basketball games. Players are segmented and classified in a system where no prior is known about the visual appearance of each team with associative embedding. A trained CNN outputs a player segmentation mask and, for each pixel, a feature vector that is similar for players belonging to the same team. Theagarajan and Bhanu \cite{Theagarajan_2021} classify soccer players by team as part of a pipeline for generating tactical performance statistics by using triplet CNNs. 

In ice hockey, Guo \textit{et al.} \cite{guo} perform team identification using the color features of hockey player uniforms. For this purpose, the uniform region (central region) of the player's bounding box is cropped. From this region, hue, saturation, and lightness (HSL) pixel values are extracted, and histograms of pixels in five essential color channels (i.e., green, yellow, blue, red, and white) are constructed. Finally, the player's team identification is determined by the channel that contains the maximum proportions of pixels. Koshkina \textit{et al.} \cite{Koshkina_2021_CVPR} use contrastive learning to classify player bounding boxes in hockey games. This self-supervised learning approach uses a CNN trained with triplet loss to learn a feature space that best separates players into two teams. Over a sequence of initial frames, they first learn two k-means cluster centres, then associate players to teams.


\section{Technical Approach}

\subsection{Player Tracking}
\subsubsection{Dataset}
\begin{figure}[t]
\begin{center}
\includegraphics[width=\linewidth, height = 3.6cm]{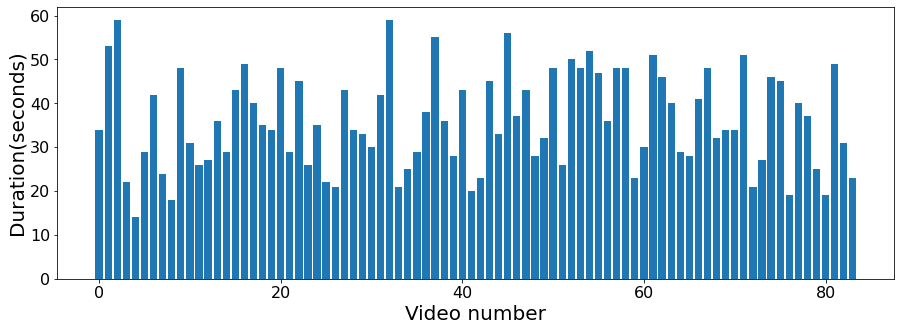}
\end{center}
  \caption{Duration of videos in the player tracking dataset. The average clip duration is $36$ seconds. }
\label{fig:clip_lengths}
\end{figure}

The player tracking dataset consists of a total of 84 broadcast NHL game clips with a frame rate of 30 frames per second (fps) and resolution of $1280 \times 720$ pixels. The average clip duration is $36$ seconds. The 84 video clips in the dataset are extracted from 25 NHL games. The duration of the clips in shown in Fig. \ref{fig:clip_lengths}. Each frame in a clip is annotated with player and referee bounding boxes and player identity consisting of player name and jersey number. The annotation is carried out with the help of open source CVAT tool\footnote{Found online at: \url{https://github.com/openvinotoolkit/cvat}}. The dataset is split such that 58 clips are used for training, 13 clips for validation, and 13 clips for testing. To prevent any game-level bias affecting the results, the split is made at the game level, such that the training clips are obtained from 17 games, validation clips from 4 games and test split from 4 games, respectively.

\subsubsection{Methodology}
 We experimented with five state-of-the-art tracking algorithms on the hockey player tracking dataset. The algorithms include four online tracking algorithms \cite{Bewley2016_sort, Wojke2017simple,zhang2020fair,tracktor_2019_ICCV} and one offline tracking algorithm \cite{Braso_2020_CVPR}. The best tracking performance (see Section \ref{section:results}) is achieved using the MOT Neural Solver tracking model \cite{Braso_2020_CVPR} re-trained on the hockey dataset. MOT Neural Solver uses the popular tracking-by-detection paradigm.
 
 
 In tracking by detection, the input is a set of object detections $O = \{o_1,.....o_n\}$, where $n$ denotes the total number of detections in all video frames. A detection $o_i$ is represented by $\{x_i,y_i,w_i,h_i, I_i,t_i\}$, where $x_i,y_i,w_i,h_i$ denotes the coordinates, width, and height of the detection bounding box. $I_i$ and $t_i$ represent the image pixels and timestamp corresponding to the detection. The goal is to find a set of trajectories $T = \{ T_1, T_2....T_m\}$ that best explains $O$ where each $T_i$ is a time-ordered set of observations. The  MOT Neural Solver models the tracking problem as an undirected graph $G=(V,E)$ , where $V=\{1,2, ..., n\}$ is the set of $n$ nodes for $n$ player detections for all video frames. In the edge set $E$, every pair of detections is connected so that trajectories with missed detections can be recovered. The problem of tracking is now posed as splitting the graph into disconnected components where each component is a trajectory $T_i$. After computing each node (detection) embedding and edge embedding using a CNN, the model then solves a graph message passing problem. The message passing algorithm classifies whether an edge between two nodes in the graph belongs to the same player trajectory. 

\subsection{Team Identification}

\subsubsection{Dataset}
\begin{figure}[t]
\begin{center}
\includegraphics[width=\linewidth]{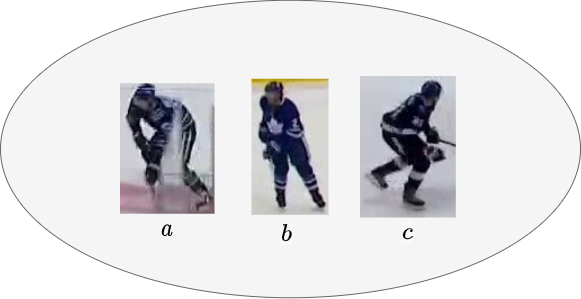}
\end{center}
  \caption{Examples of `blue' class in the team identification dataset. Home jersey of teams such as (a) Vancouver Canucks (b) Toronto Maple Leafs and (c) Tampa Bay Lightning are blue in appearance and hence are put in the same class.}
\label{fig:team_examples}
\end{figure}

\begin{figure}[t]
\begin{center}
\includegraphics[width=\linewidth]{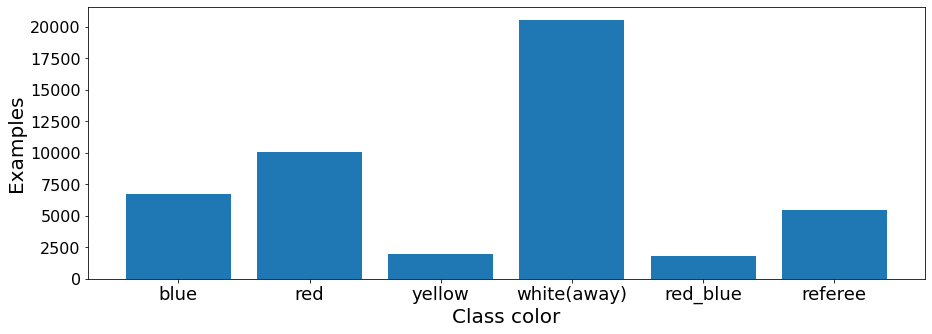}
\end{center}
  \caption{Classes in team identification and their distribution. The `ref' class denotes referees.}
\label{fig:team_data_dist}
\end{figure}

The team identification dataset is obtained from the same games and clips used in the player tracking dataset. The train/validation/test splits are also identical to player tracking data. We take advantage of the fact that the away team in NHL games usually wear a predominantly white colored jersey with color stripes and patches, and the home team wears a dark colored jersey. For example, the Toronto Maple Leafs and the Tampa Bay Lightning both have dark blue home jerseys and therefore can be put into a single `Blue' class. We therefore build a dataset with five classes (blue, red, yellow, white, red-blue and referees) with each class composed of images with same dominant color.  The data-class distribution is shown in Fig. \ref{fig:team_data_dist}. Fig. \ref{fig:team_examples} shows an example of the blue class from the dataset. The training set consists of $32419$ images. The validation and testing set contain $6292$ and $7898$ images respectively.

\subsubsection{Methodology}

For team identification, we use a ResNet18 \cite{resnet} pretrained on the ImageNet dataset \cite{deng2009imagenet}, and train the network on the team identification dataset by replacing the final fully connected layer to output six classes. The images are scaled to a resolution of $224 \times 224$ pixels for training. During inference, the network classifies whether a bounding box belongs to the away team (white color), the home team (dark color), or the referee class. For inferring the team for a player tracklet, the team identification model is applied to each image of the tracklet and a simple majority vote is used to assign a team to the tracklet. This way, the tracking algorithm helps team identification by resolving errors in team identification.

\subsubsection{Training Details}

We use the Adam optimizer with an initial learning rate of .001 and a weight decay of .001 for optimization. The learning rate is reduced by a factor of $\frac{1}{3}$  at regular intervals during the training process.  We do not perform data augmentation since performing color augmentation on white away jerseys makes it resemble colored home jerseys.

\begin{table}[!t]
    \centering 
    \caption{Network architecture for the player identification model. k, s, d and p denote kernel dimension, stride, dilation size and padding respectively. $Ch_{i}$, $Ch_{o}$ and $b$ denote the number of channels going into and out of a block, and batch size, respectively.}
    \footnotesize
    \setlength{\tabcolsep}{0.2cm}
    \begin{tabular}{c}
    \hline \textbf{Input: Player tracklet} $b\times 30 \times 3 \times 300\times300$  \\\hline\hline
    
    \textbf{ResNet18 backbone} \\ \hline
     \textbf{Layer 1}:
     Conv1D \\
 $Ch_{i} = 512, Ch_{o} = 512$ \\
   (k = $3$,
      s = $3$,
   p = $0$,
   d = $1$) \\
      Batch Norm 1D    \\
     ReLU  \\ \hline
          \textbf{Layer 2}:
     Conv1D \\
 $Ch_{i} = 512, Ch_{o} = 512$ \\
   (k = $3$,
      s = $3$,
   p = $1$,
   d = $1$) \\
      Batch Norm 1D    \\
     ReLU  \\
 \hline
 \textbf{Layer 3}:
 Conv2D \\
 $Ch_{i} = 512, Ch_{o} = 128$ \\
   (k = $3$,
      s = $1$,
   p = $0$,
   d = $1$) \\
      Batch Norm 1D    \\
     ReLU 
    \\ \hline
    \textbf{Layer 4}:
    Fully connected\\
   $Ch_{i} = 128, Ch_{o} = 86$ \\ \hline
        \textbf{Output} $b\times86$ \\ \hline
    \end{tabular}
    \label{table:player_branch_network}
\end{table}

\subsection{Player Identification}

\subsubsection{Image Dataset}
\label{subsubsection:image_dataset}
 The player identification image dataset \cite{Vats2021MultitaskLF} consists of $54,251$ player bounding boxes obtained from 25 NHL games. The NHL game video frames are of resolution $1280 \times 720$ pixels. The dataset contains $81$ jersey number classes, including an additional $null$ class for no jersey number visible. The player head and bottom of the images are cropped such that only the jersey number (player torso) is visible. Images from $17$ games are used for training, four games for validation and four games for testing. The dataset is highly imbalanced such that the ratio between the most frequent and least frequent class is $92$. The dataset covers a range of real-game scenarios such as occlusions, motion blur and self-occlusions.

\begin{figure*}[t]
\begin{center}
\includegraphics[width=\linewidth]{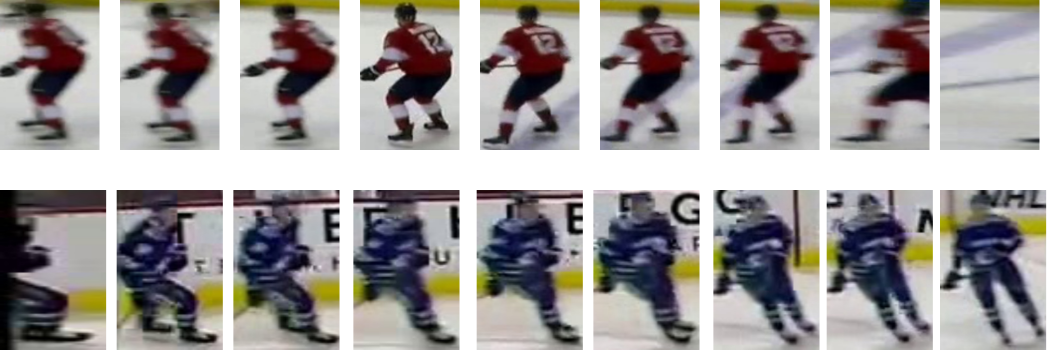}
\end{center}
  \caption{Examples of two tracklets in the player identification dataset. \textbf{Top row:} Tracklet represents a case when the jersey number 12 is visible in only a subset of frames. \textbf{Bottom row:} Example when the jersey number is never visible over the whole tracklet. }
\label{fig:tracklet_data}
\end{figure*}

\subsubsection{Tracklet Dataset}

\begin{figure}[t]
\begin{center}
\includegraphics[width=\linewidth, height = 3.5cm]{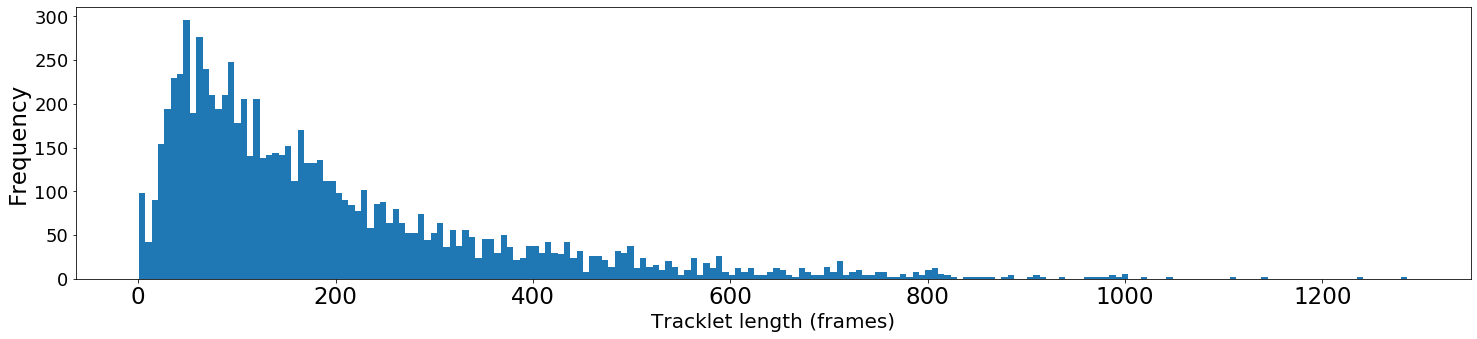}
\end{center}
  \caption{Distribution of tracklet lengths in frames of the player identification dataset. The distribution is positively skewed with the average length of a player tracklet as $191$ frames.}
\label{fig:tracklet_dist}
\end{figure}

\begin{figure*}[t]
\begin{center}
\includegraphics[width=\linewidth]{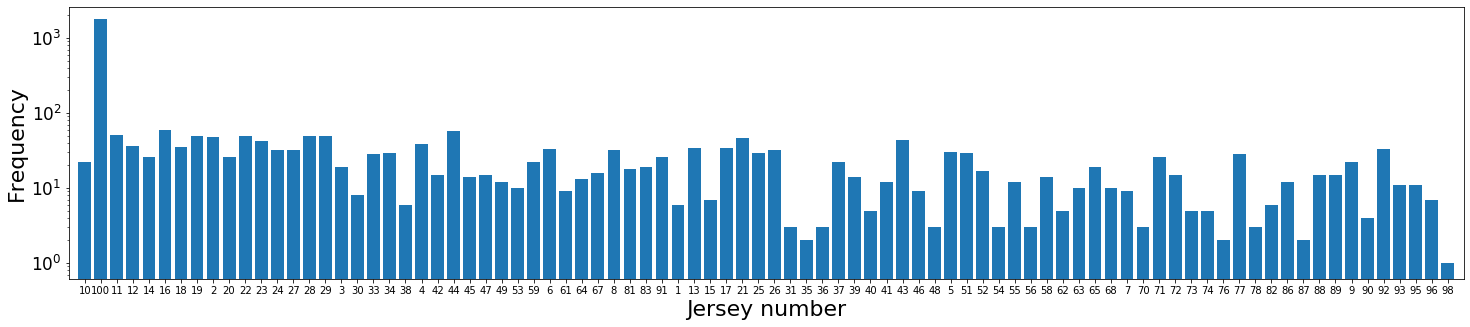}
\end{center}
  \caption{Class distribution in the player tracklet identification dataset.  The dataset is heavily imbalanced with the $null$ class (denoted by class $100$) consisting of $50.4\%$ of tracklet examples.}
\label{fig:class_jersey}
\end{figure*}

The player identification tracklet dataset consists of $3510$ player tracklets. The tracklet bounding boxes and identities are annotated manually. The manually annotated tracklets simulate the output of a tracking algorithm. The tracklet length distribution is shown in Fig. \ref{fig:tracklet_dist}. The average length of a player tracklet is $191$ frames (6.37 seconds in a 30 frame per second video). It is important to note that the player jersey number is visible in only a subset of tracklet frames. Fig. \ref{fig:tracklet_data} illustrates two tracklet examples from the dataset. The dataset is divided into 86 jersey number classes with one $null$ class representing no jersey number visible. The class distribution is shown in Fig. \ref{fig:class_jersey}. The dataset is heavily imbalanced with the $null$ class consisting of $50.4\%$ of tracklet examples. The training set contains $2829$ tracklets, $176$ validation tracklets and $505$ test  tracklets. The game-wise training/testing data split is identical in all the four datasets discussed.

\subsubsection{Network Architecture}
\label{subsection:network_architecture}
Let $T = \{ o_1,o_2....o_n\}$ denote a player tracklet where each $o_i$ represents a player bounding box. The player head and bottom in the bounding box $o_i$ are cropped such that only the jersey number is visible. Each resized image $I_{i} \in \mathbb{R}^{300\times\ 300\times3}$ corresponding to the bounding box $o_i$ is input into a backbone 2D CNN, which outputs a set of time-ordered features $\{F = \{f_1,f_2.....f_n\}, f_i \in \mathbb{R}^{512} \}$. The features $F$ are input into a 1D temporal convolutional network that outputs probability $p\in \mathbb{R}^{86}$ of the tracklet belonging to a particular jersey number class. The architecture of the 1D CNN is shown in Fig. \ref{fig:jn_arch}.\par 
The network consists of a ResNet18 \cite{resnet} based 2D CNN backbone pretrained on the player identification image dataset (Section \ref{subsubsection:image_dataset}). The weights of the ResNet18 backbone network are kept frozen while training. The 2D CNN backbone is followed by three 1D convolutional blocks each consisting of a 1D convolutional layer, batch normalization, and ReLU activation. Each block has a kernel size of three and dilation of one. The first two blocks have a larger stride of three, so that the initial layers have a larger receptive field to take advantage of a large temporal context. Residual skip connections are added to aid learning. The exact architecture is shown in Table \ref{table:player_branch_network}. Finally, the activations obtained are pooled using mean pooling and passed through a fully connected layer with 128 units. The logits obtained are softmaxed to obtain jersey number probabilities. Note that the model accepts fixed length training sequences of length $n=30$ frames as input, but the training tracklets are hundreds of frames in length (Fig. \ref{fig:tracklet_dist}). Therefore, $n=30$ tracklet frames are sampled with a random starting frame from the training tracklet. This serves as a form of data augmentation since at every training iteration, the network processes a randomly sampled set of frames from an input tracklet.

\subsubsection{Training Details}

 To address the severe class imbalance present in the tracklet dataset, the tracklets are sampled intelligently such that the $null$ class is sampled with a probability $p_0=0.1$. The network is trained with the help of cross entropy loss. We use Adam optimizer for training with a initial learning rate of $.001$ with a batch size of $15$. The learning rate is reduced by a factor of $\frac{1}{5}$ after iteration numbers 2500, 5000, and 7500. Several data augmentation techniques such as random cropping, color jittering, and random rotation are also used. All experiments are performed on two Nvidia P-100 GPUs.

\subsubsection{Inference}
\label{subsubsection:inference}
During inference, we need to assign a single jersey number label to a test tracklet of $k$ bounding boxes $T_{test} = \{ o_1,o_2....o_k\}$. Here $k$ can be much greater than $n = 30$. So, a sliding window technique is used where the network is applied to the whole test tracklet $T_{test}$ with a stride of one frame to obtain window probabilities $P = \{p_1,p_2,...p_k\}$ with each $p_i \in R^{86}$. The probabilities $P$ are aggregated to assign a single jersey number class to a tracklet. To aggregate the probabilities $P$, we filter out the tracklets where the jersey number is visible. To do this we first train a ResNet18 classifier $C^{im}$ (same as the backbone of discussed in Section \ref{subsection:network_architecture}) on the player identification image dataset. The classifier $C^{im}$ is run on every image of the tracklet. A jersey number is assumed to be absent on a tracklet if the probability of the absence of jersey number $C^{im}_{null}$ is greater than a threshold $\theta$ for each image in the tracklet. The threshold $\theta$ is determined using the player identification validation set. The tracklets for which the jersey number is visible, the probabilities are averaged to obtain a single probability vector $p_{jn}$, which represents the probability distribution of the jersey number in the test tracklet $T_{test}$. For post-processing, only those probability vectors $p_i$ are averaged for which $argmax(p_i) \ne null$. 


The rationale behind using visibility filtering and post-processing step is that a large tracklet with hundreds of frames may have the number visible in only a few frames and therefore, a simple averaging of probabilities $P$ will often output $null$. The proposed inference technique allows the network to ignore the window probabilities corresponding to the $null$ class if a number is visible in the tracklet. The whole algorithm is illustrated in Algorithm \ref{algorithm:inference}.



\begin{algorithm}[]
\setstretch{1}
\SetAlgoLined
\textbf{Input}: Tracklet $T_{test}= \{ o_1,o_2....o_k\}$, image-wise jersey number classifier $C^{im}$, Tracklet id model  $\mathcal{P}$, Jersey number visibility threshold $\theta$  \\
\textbf{Output}: Identity $Id$, $p_{jn}$ \\
\textbf{Initialize}: $vis = false$ \\

$P = \mathcal{P}(T_{test})$    \tcp*[h]{using sliding window}

\For{$o_i \quad in \quad T_{test} $}{
\If{$ C^{im}_{null}(o_i) < \theta $}{
    $vis = true$\\
    $break$
} 

}
\If{$vis == true$}{

$P^\prime = \{p_i \in P : argmax(p_i) \ne null\} $    \tcp*[h]{post-processing} \\
$p_{jn} = mean(P^\prime)$\\
$Id = argmax(p_{jn})$
}
\Else{
$Id = null$

}

 \caption{Algorithm for inference on a tracklet.}
 \label{algorithm:inference}
\end{algorithm}

\begin{table*}[!t]

    \centering
    \caption{Tracking performance of MOT Neural Solver model for the 13 test videos ($\downarrow$ means lower is better, $\uparrow$ means higher is better). }
    \footnotesize
    \setlength{\tabcolsep}{0.2cm}
    \begin{tabular}{c|c|c|c|c|c}\hline
  
       Video number & IDF1$\uparrow$ & MOTA $\uparrow$& ID-switches $\downarrow$ & False positives (FP)$\downarrow$ & False negatives (FN) $\downarrow$\\\hline
     
     1  & $78.53$ &  $94.95$ & $23$ & $100$ & $269$\\
      2  & $61.49$ &  $93.29$ & $26$ & $48$ & $519$\\
       3  & $55.83$ &  $95.85$ & $43$ & $197$ & $189$\\
        4  & $67.22$ &  $95.50$ & $31$ & $77$ & $501$\\
    5   & $72.60$ &  $91.42$ & $40$ & $222$ & $510$\\
      6  & $66.66$ &  $90.93$ & $38$ & $301$ & $419$\\
      7  & $49.02$ &  $94.89$ & $59$ & $125$ & $465$\\
       8  & $50.06$ &  $92.02$ & $31$ & $267$ & $220$\\
        9  & $53.33$ &  $96.67$ & $30$ & $48$ & $128$\\
    10   & $55.91$ &  $95.30$ & $26$ & $65$ & $193$\\
      11  & $56.52$ &  $96.03$ & $40$ & $31$ & $477$\\
        12  & $87.41$ &  $94.98$ & $14$ & $141$ & $252$\\
    13   & $62.98$ &  $94.77$ & $30$ & $31$ & $252$\\
   
    \end{tabular}
    \label{table:video_wise_track}
\end{table*}

\begin{table*}[!t]

    \centering
    \caption{Comparison of the overall tracking performance on test videos the hockey player tracking dataset. ($\downarrow$ means lower is better, $\uparrow$ mean higher is better) }
    \footnotesize
    \setlength{\tabcolsep}{0.2cm}
    \begin{tabular}{c|c|c|c|c|c}\hline
  
       Method & IDF1$\uparrow$ & MOTA $\uparrow$& ID-switches $\downarrow$ & False positives (FP)$\downarrow$ & False negatives (FN) $\downarrow$\\\hline
     
     SORT \cite{Bewley2016_sort} & $53.7$ &  $92.4$ & $673$ & $2403$ & $5826$\\
      Deep SORT \cite{Wojke2017simple} & $59.3$ &  $94.2$ & $528$ & $1881$ & $4334$\\
       Tracktor \cite{tracktor_2019_ICCV} & $56.5$ &  $94.4$ & $687$ & $1706$ & $4216$\\
        FairMOT \cite{zhang2020fair} & $61.5$ &  $91.9$ & $768$ & $1179$ & $7568$\\
        MOT Neural Solver \cite{Braso_2020_CVPR}  & $\textbf{62.9}$ &  $\textbf{94.5}$ & $\textbf{431}$ & $1653$ & $4394$\\
   
    \end{tabular}
    \label{table:tracking_results}
\end{table*}

\subsection{Overall System}
\label{subsection:pipeline}
The player tracking, team identification, and player identification methods discussed are combined together for tracking and identifying players and referees in broadcast video shots. Given a test video shot, we first run player detection and  tracking to obtain a set of player tracklets $\tau = \{ T_1, T_2,....T_n \}$. For each tracklet $T_i$ obtained, we run the player identification model to obtain the player identity. We take advantage of the fact that the player roster is available for NHL games through play-by-play data, hence we can focus only on players actually present in the team. To do this, we construct vectors $v_a$ and $v_h$ that contain information about which jersey numbers are present in the away and home teams, respectively. We refer to the vectors $v_h$ and $v_a$ as the \textit{roster vectors}. Assuming we know the home and away roster, let $H$ be the set of jersey numbers present in the home team and $A$ be the set of jersey numbers present in away team. Let $p_{jn} \in \mathbb{R}^{86}$ denote the probability of the jersey number present in the tracklet.  Let $null$ denote the no-jersey number class and $j$ denote the index associated with jersey number $\eta_j$ in $p_{jn}$ vector. 

\begin{align}
    v_h[j] &=1, \text{if} \; \eta_j \in H\cup\{null\} \\
    v_h[j] &= 0, \; otherwise 
\end{align}

\noindent similarly,  

\begin{align}
    v_a[j] &=1, \text{if} \; \eta_j \in A\cup\{null\} \\
    v_a[j] &= 0, \; otherwise 
\end{align}

We multiply the probability scores $p_{jn} \in R^{86}$ obtained from the player identification by  $v_h \in R^{86}$ if the player belongs to home team or $v_a \in R^{86}$ if the player belongs to the away team. The determination of player team is done through the trained team identification model. The player identity $I$ is determined through \begin{equation}Id=argmax(p_{jn} \odot v_h) \end{equation} (where $\odot$ denotes element-wise multiplication) if the player belongs to home team, and \begin{equation}Id=argmax(p_{jn} \odot v_a)\end{equation} if the player belongs to the away team. The overall algorithm is summarized in Algorithm \ref{algorithm:holistic}. Fig. \ref{fig:pipeline} depicts the overall system.

\begin{algorithm}[]
\setstretch{1}
\SetAlgoLined
\textbf{Input}: Input Video $V$, Tracking model $T_r$ , Team ID model $\mathcal{T}$, Player ID model $\mathcal{P}$, $v_h$ , $v_a$\\
\textbf{Output}: Identities $\mathcal{ID} = \{Id_1, Id_2.....Id_n\}$\\
\textbf{Initialize}: $\mathcal{ID} = \phi$\\
$\tau = \{ T_1, T_2,....T_n \} = T_r(V)$\\
\For{$T_i \quad in \quad \tau $}
{
$team = \mathcal{T}(T_i)$\\
 $p_{jn} = \mathcal{P}(T_i)$\\
\uIf{$team == home$}{
   
     $Id=argmax(p_{jn} \odot v_h)$ 
    
}
\uElseIf{$team == away$}{

 $Id=argmax(p_{jn} \odot v_a)$ 
}
\Else{
$Id = ref$
}
$\mathcal{ID} = \mathcal{ID} \cup Id $

}

 \caption{Holistic algorithm for player tracking and identification.}
 \label{algorithm:holistic}
\end{algorithm}

\section{Results}
\label{section:results}

\subsection{Player Tracking}

The MOT Neural Solver algorithm is compared with four state of the art algorithms for tracking. The methods compared to are Tracktor \cite{tracktor_2019_ICCV}, FairMOT \cite{zhang2020fair}, Deep SORT \cite{Wojke2017simple} and SORT \cite{Bewley2016_sort}. Player detection is performed using a Faster-RCNN network \cite{NIPS2015_14bfa6bb} with a ResNet50 based Feature Pyramid Network (FPN) backbone \cite{fpn} pre-trained on the COCO dataset \cite{coco} and fine tuned on the hockey tracking dataset. The object detector obtains an average precision (AP) of $70.2$ on the test videos (Table \ref{table:player_det_results}). The accuracy metrics for tracking used are the CLEAR MOT metrics \cite{mot} and Identification F1 score (IDF1) \cite{idf1}. An important metric is the number of identity switches (IDSW), which occurs when a ground truth ID $i$ is assigned a tracked ID $j$ when the last known assignment ID was $k \ne j$. A low number of identity switches is an indicator of accurate tracking performance. For sports player tracking, the IDF1 is considered a better accuracy measure than Multi Object Tracking accuracy (MOTA) since it measures how consistently the identity of a tracked object is preserved with respect to the ground truth identity. The overall  results are shown if Table \ref{table:tracking_results}. The MOT Neural Solver model obtains the highest MOTA score of $94.5$ and IDF1 score of $62.9$ on the test videos.

\subsubsection{Analysis}

From Table \ref{table:tracking_results} it can be seen that the MOTA score of all methods is above $90\%$. This is because MOTA is calculated as
\begin{equation}
\label{MOT accuracy equation}
    MOTA = 1 - \frac{\Sigma_{t}(FN_{t} + FP_{t} + IDSW_{t})}{\Sigma_t GT_{t}}
\end{equation} 
where $t$ is the frame index and $GT$ is the number of ground truth objects. MOTA metric counts detection errors through the sum $FP+FN$ and association errors through $IDSWs$. Since false positives (FP) and false negatives (FN) heavily rely on the performance of the player detector, the MOTA metric highly depends on the performance of the detector. For hockey player tracking, the player detection accuracy is high  because of sufficiently large size of players in broadcast video and a reasonable number of players and referees (with a fixed upper limit) to detect in the frame. Therefore, the MOTA score for all methods is high.\par
The MOT Neural Solver method achieves the highest IDF1 score of $62.9$ and significantly lower identity switches than the other methods. This is because pedestrian trackers use a linear motion model assumption which does not perform well with motion of hockey players. Sharp changes in player motion often leads to identity switches. The MOT Neural Solver model, in contrast, has no such assumptions since it poses tracking as a graph edge classification problem.  \par
Table \ref{table:video_wise_track} shows the performance of the MOT Neural solver for each of the 13 test videos. We do a failure analysis to determine the cause of identity switches and low IDF1 score in some videos. The major sources of identity switches are severe occlusions and player going out of field of view (due to camera panning and/or player movement). We define a pan-identity switch as an identity switch resulting from a player leaving and re-entering camera field of view due to panning. It is very difficult for the tracking model to maintain identity in these situations since players of the same team look identical and a player going out of the camera field of view at a particular point in screen coordinates can re-enter at any other point. We try to estimate the proportion of pan-identity switches to determine the contribution of panning to total identity switches. \par

\begin{figure}[t]
\begin{center}
\includegraphics[width=\linewidth]{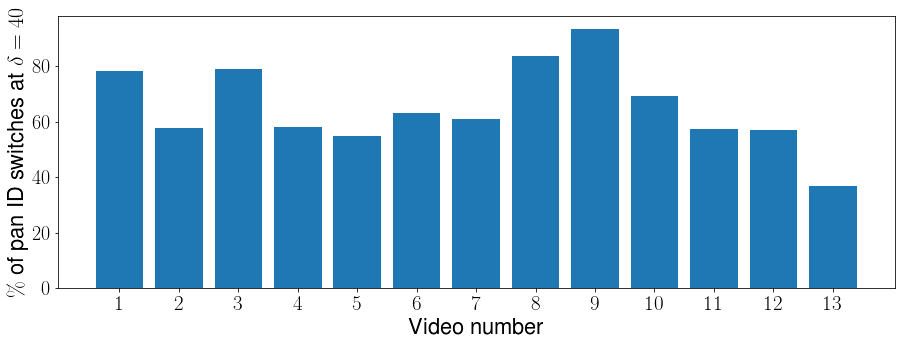}
\end{center}
  \caption{Proportion of pan-identity switches for all videos at a threshold of $\delta = 40$ frames. On average, pan-identity switches account for $65\%$ of identity switches. }
\label{fig:all_pidsw}
\end{figure}

\begin{figure}[t]
\begin{center}
\includegraphics[width=\linewidth]{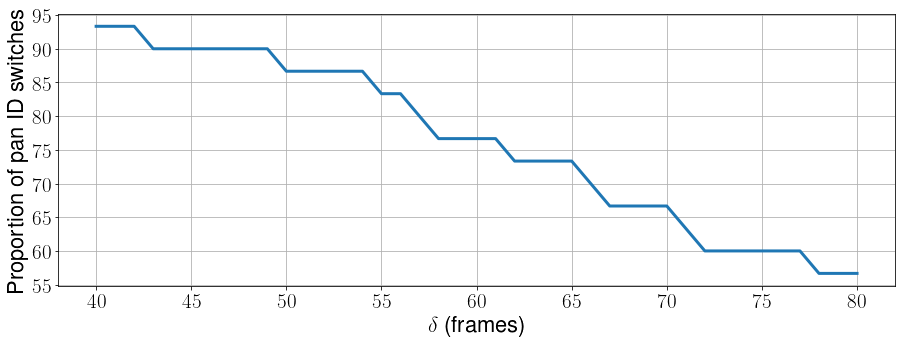}
\end{center}
  \caption{Proportion of pan identity switches vs. $\delta$ plot for video number $9$. Majority of the identity switches ($~90\%$ at a threshold of $\delta= 40$ frames) occur due to camera panning, which is the main cause of error.}
\label{fig:vid_9_pidsw}
\end{figure}

To estimate the number of pan-identity switches, since we have quality annotations, we make the assumption that the ground truth annotations are accurate and there are no missing annotations in ground truth. Based on this assumption, there is a significant time gap between two consecutive annotated detections of a player only when the player leaves the camera field of view and comes back again. Let  $T_{gt} = \{o_1, o_2,..., o_n \}$ represent a ground truth tracklet, where  $o_i = \{x_i,y_i,w_i,h_t,I_i, t_i\}$ represents a ground truth detection. A pan-identity switch is expected to occur during tracking when the difference between timestamps (in frames) of two consecutive ground truth detections $i$ and $j$ is greater than a sufficiently large threshold $\delta$.  That is \begin{equation}
    (t_i - t_j) > \delta 
\end{equation}
Therefore, the total number of pan-identity switches in a video is approximately calculated as
\begin{equation}
    \sum_G \mathbb{1}( t_i - t_j > \delta ) 
\end{equation}
where the summation is carried out over all ground truth trajectories and $\mathbb{1}$ is an indicator function. Consider the video number $9$ having $30$ identity switches and an IDF1 of $53.33$. We plot the proportion of pan identity switches (Fig \ref{fig:vid_9_pidsw}), that is \begin{equation}
     = \frac{\sum_G \mathbb{1}( t_i - t_j > \delta )}{IDSWs} 
\end{equation}
against $\delta$, where $\delta$ varies between $40$ and $80$ frames. In video number $9$ video $IDSWs = 30$.  From Fig. \ref{fig:vid_9_pidsw} it can be seen that majority of the identity switches ($~90\%$ at a threshold of $\delta= 40$ frames) occur due to camera panning, which is the main cause of error. Visually investigating the video confirmed the statement. Fig. \ref{fig:all_pidsw} shows the proportion of pan-identity switches for all videos at a threshold of $\delta = 40$ frames. On average, pan identity switches account for $65\%$ of identity switches in the videos. This shows that the tracking model is able to tackle occlusions and lack of detections with the exception of extremely cluttered scenes.

 \begin{table}[!t]

    \centering
    \caption{Player detection results on the test videos. $AP$ stands for Average Precision. $AP_{50}$ and $AP_{75}$ are the average precision at an Intersection over Union (IoU) of 0.5 and 0.75 respectively.}
    \footnotesize
    \setlength{\tabcolsep}{0.2cm}
    \begin{tabular}{c|c|c}\hline
  
         $AP$ & $AP_{50}$ & $AP_{75}$   \\\hline
     
    $70.2$ &  $95.9$    & $87.5$ 

    \end{tabular}
    \label{table:player_det_results}
\end{table} 

\subsection{ Team Identification}

\begin{figure*}[t]
\begin{center}
\includegraphics[width=\linewidth]{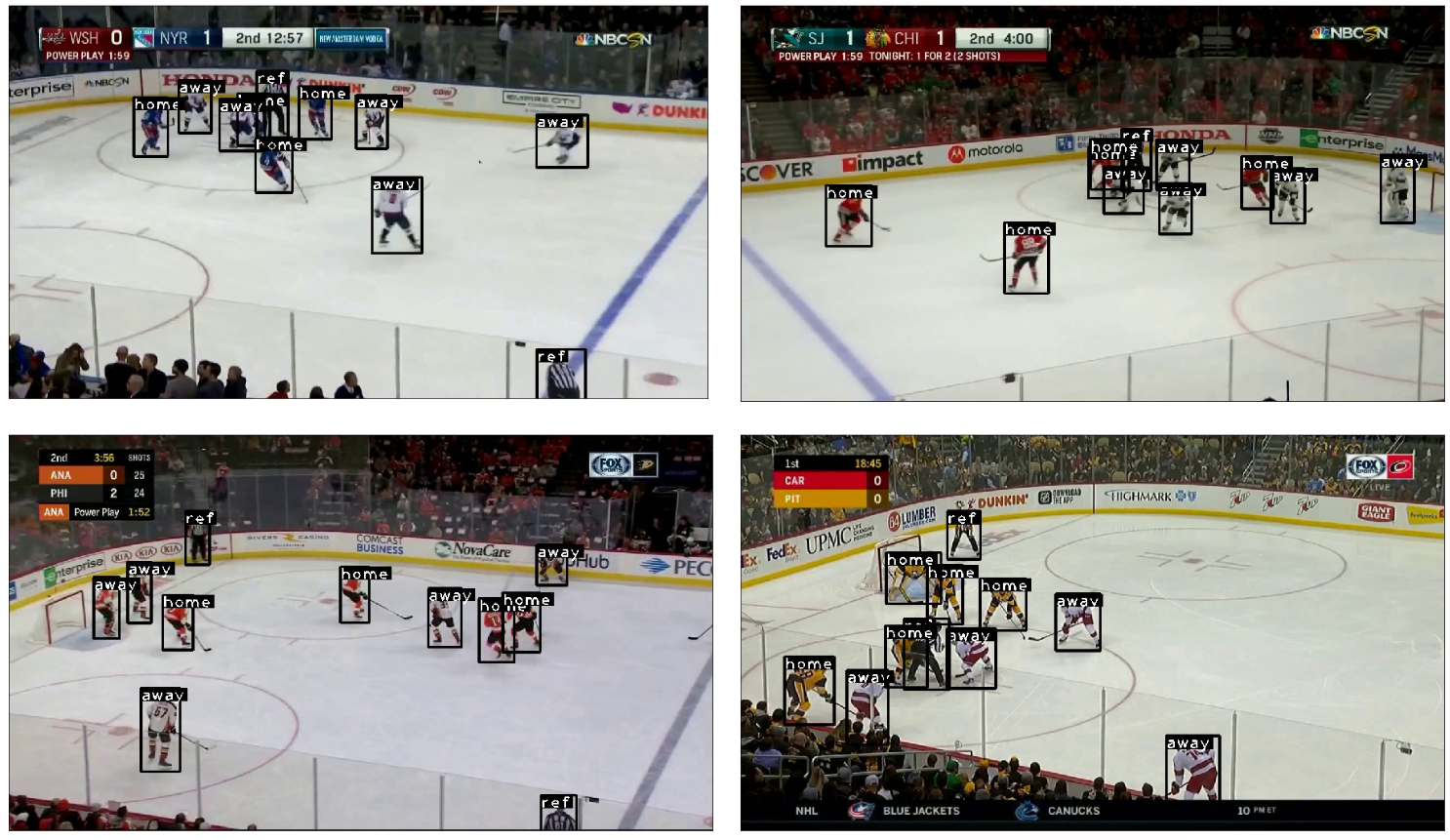}
\end{center}
  \caption{Team identification results from four different games that are each not present in the team identification dataset. The model performs well on data not present in dataset, which demonstrates the ability to generalize well on out of sample data points. }
\label{fig:team_results}
\end{figure*}

The team identification model obtains an accuracy of $96.6\%$ on the team identification test set. Table \ref{table:team_id_results} shows the macro averaged precision, recall and F1 score for the results. The model is also able to correctly classify teams in the test set that are not present in the training set. Fig. \ref{fig:team_results} shows some qualitative results where the network is able to generalize on videos absent in training/testing data. We compare the model to color histogram features as a baseline. Each image in the dataset was cropped such that only the upper half of jersey is visible. A color histogram was obtained from the RGB representation of each image, with $n_{bins}$ bins per image channel. Finally a support vector machine (SVM) with an radial basis function (RBF) kernel was trained on the normalized histogram features. The optimal SVM hyperparameters and number of histogram bins were  determined using grid search by doing a five-fold cross-validation on the combination of training and validation set. The optimal hyperparameters obtained were $C=10$ , $\gamma = .01$ and $n_{bins}=12$. Compared to the SVM model, the deep network approach performs $14.6\%$ better on the test set  demonstrating that the deep network (CNN) based approach is superior to simple handcrafted color histogram features.

\subsection{Player Identification}

\begin{figure}[t]
\begin{center}
\includegraphics[width=\linewidth]{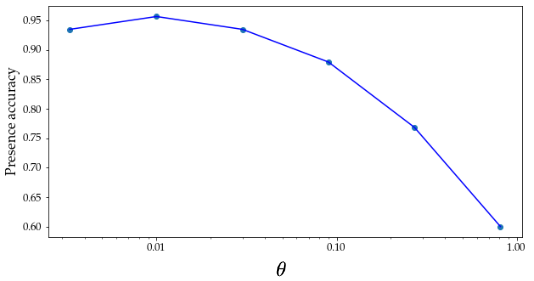}
\end{center}
  \caption{Jersey number presence accuracy vs. $\theta$ (threshold for filtering out tracklets where jersey number is not visible) on the validation set.  The values of $\theta$ tested are $\theta = \{0.0033, 0.01, 0.03, 0.09, 0.27, 0.81\}$. The highest accuracy is attained at $\theta=0.01$. }
\label{fig:theta_presence}
\end{figure}

 \begin{table}[!t]

    \centering
    \caption{Team identification accuracy on the team-identification test set.}
    \footnotesize
    \setlength{\tabcolsep}{0.2cm}
    \begin{tabular}{c|c|c|c|c}\hline
  
       Method & Accuracy & Precision & Recall & F1 score  \\\hline
     
  Proposed &  $\mathbf{96.6}$ &  $\mathbf{97.0}$    & $\mathbf{96.5}$ & $\mathbf{96.7}$ \\
    SVM with color histogram & $82.0$ & $81.7$  & $81.5$  & $81.5$   \\

    \end{tabular}
    \label{table:team_id_results}
\end{table}

 \begin{table*}[!t]

    \centering
    \caption{Ablation study on different methods of aggregating probabilities for tracklet confidence scores.}
    \footnotesize
    \setlength{\tabcolsep}{0.2cm}
    \begin{tabular}{c|c|c|c|c}\hline
  
       Method & Accuracy & F1 score & Visiblility filtering & Postprocessing\\\hline
     
     Majority voting  & $80.59 \%$ &  $80.40\%$   & \checkmark & \checkmark \\
     Probability averaging  & $75.64\%$ &  $75.07\%$  &  &  \\

     Proposed w/o postprocessing  & $80.80\%$ &  $79.12\%$ & \checkmark & \\
        Proposed w/o visibility filtering  & $50.10\%$ &  $48.00\%$ &  & \checkmark\\
         Proposed  & $\textbf{83.17\%}$ &  $\textbf{83.19\%}$ & \checkmark & \checkmark\\

    \end{tabular}
    \label{table:logit_averaging}
\end{table*}

\begin{figure*}[t]
\begin{center}
\includegraphics[width=\linewidth, height=3.5cm]{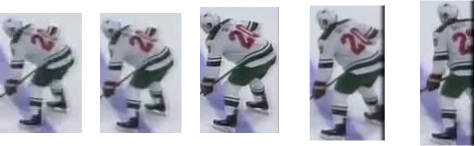}
\end{center}
  \caption{Some frames from a tracklet where the model is able to identify the number 20 where 0 is at a tilted angle in majority of bounding boxes.  }
\label{fig:success}
\end{figure*}

\begin{figure*}[t]
\begin{center}
\includegraphics[width=\linewidth, height=3.5cm]{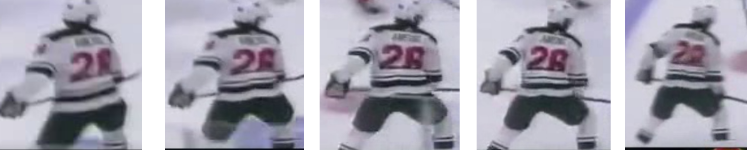}
\end{center}
  \caption{Some frames from a tracklet where $6$ appears as $8$ due to motion blur and folds in the player jersey leading to error in classification. }
\label{fig:26_28}
\end{figure*}

\label{subsection:player_identification}

The proposed player identification network attains an accuracy of $83.17 \%$ on the test set. We compare the network with Chan \textit{et al.} \cite{CHAN2021113891} who use a secondary CNN model for aggregating probabilities on top of an CNN+LSTM model. Our proposed inference scheme, on the contrary, does not require any additional network. Since the code and dataset for Chan \textit{et al.} \cite{CHAN2021113891} is not publicly available, we re-implemented the model by scratch and trained and evaluated the model on our dataset. The proposed network performs $9.9\%$ better than Chan \textit{et al.} \cite{CHAN2021113891}. The network proposed by Chan \textit{et al.} \cite{CHAN2021113891} processes shorter sequences of length 16 during training and testing, and therefore exploits less temporal context than the proposed model with sequence length 30. Also, the secondary CNN used by Chan \textit{et al.} \cite{CHAN2021113891} for aggregating tracklet probability scores easily overfits on our dataset. Adding $L2$ regularization while training the secondary CNN  proposed in Chan \textit{et al.}  \cite{CHAN2021113891} on our dataset also did not improve the performance. This is  because our dataset is half the size and is more skewed than the one used in Chan \textit{et al.}  \cite{CHAN2021113891}, with the $null$ class consisting of half the examples in our case.  The higher accuracy indicates that the proposed network and training methodology involving intelligent sampling of the $null$ class and the proposed inference scheme works better on our dataset. Additionally, temporal 1D CNNs have been reported to perform better than LSTMs in handling long range dependencies \cite{BaiTCN2018}, which is verified by the results.\par

The network is able to identify digits during motion blur and unusual angles (Fig. \ref{fig:success}). Upon inspecting the error cases, it is seen that when a two digit jersey number is misclassified, the predicted number and ground truth often share one digit. This phenomenon is observed in $85\%$ of misclassified two digit jersey numbers. For example, 55 is misclassified as 65 and 26 is misclassified as 28 since 6 often looks like 8 (Fig. \ref{fig:26_28}) because of occlusions and folds in player jerseys.

 \begin{table*}[!t]

    \centering
    \caption{Ablation study on different kinds of data augmentations applied during training. Removing any one of the applied augmentation techniques decreases the overall accuracy and F1 score.}
    \footnotesize
    \setlength{\tabcolsep}{0.2cm}
    \begin{tabular}{c|c|c|c|c}\hline
  
       Accuracy & F1 score & Color & Rotation  & Random cropping \\\hline
     
  $\textbf{83.17\%}$ &  $\textbf{83.19\%}$ &  \checkmark    & \checkmark & \checkmark \\
    $81.58\%$ & $82.00\%$ & \checkmark   & \checkmark  &  \\

     $81.58\%$ &   $81.64\%$   & \checkmark   &  & \checkmark\\
        $81.00\%$  & $81.87\%$  &   & \checkmark  & \checkmark\\

    \end{tabular}
    \label{table:augmentation}
\end{table*}

\begin{figure*}[t]
\begin{center}
\includegraphics[width=\linewidth]{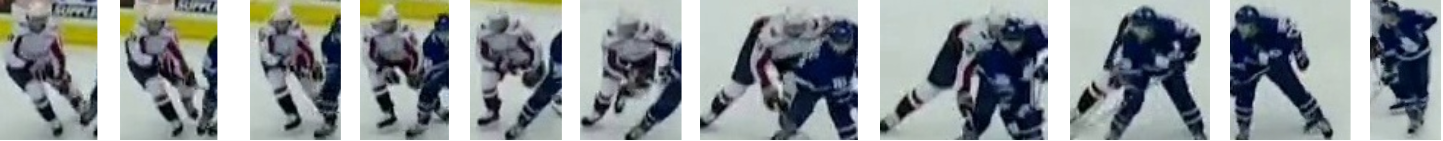}
\end{center}
  \caption{Example of a tracklet where the same identity is assigned to two different players due to an identity switch. This kind of errors in player tracking gets carried over to player identification, since a single jersey number cannot be associated with this tracklet.}
\label{fig:id_switch}
\end{figure*}

The value of $\theta$ (threshold for filtering out tracklets where jersey number is not visible) is determined using the validation set. In Fig \ref{fig:theta_presence}, we plot the percentage of validation tracklets correctly classified for presence of jersey number versus the parameter $\theta$. The values of $\theta$ tested are $\theta = \{0.0033, 0.01, 0.03, 0.09, 0.27, 0.81\}$. The highest accuracy of $95.64\%$ at $\theta = 0.01$.  A higher value of $\theta$ results in more false positives for jersey number presence. A $\theta$ lower than $0.01$ results in more false negatives. We therefore use the value of $\theta=0.01$ for doing inference on the test set. \par

\subsubsection{Ablation studies} 
We perform ablation studies in order to study how data augmentation and  inference techniques affect the player identification network performance.


\textbf{Data augmentation}: We perform several data augmentation techniques to boost player identification performance such data color jittering, random cropping, and random rotation by rotating each image in a tracklet by $\pm 10$ degrees. Note that since we are dealing with temporal data, these augmentation techniques are applied per tracklet instead of per tracklet-image. In this section, we investigate the contribution of each augmentation technique to the overall accuracy. Table \ref{table:augmentation} shows the accuracy and weighted macro F1 score values after removing these augmentation techniques. It is observed that removing any one of the applied augmentation techniques decreases the overall accuracy and F1 score.

\textbf{Inference technique}:
We perform an ablation study to determine how our tracklet score aggregation scheme of averaging probabilities after filtering out tracklets based on jersey number presence compares with other techniques. Recall from section \ref{subsubsection:inference} that for inference, we perform \textit{visibility filtering} of tracklets and evaluate the model only on tracklets where jersey number is visible.   We also include a \textit{post-processing} step where only those window  probability vectors $p_i$ are averaged for which $argmax(p_i) \ne null$. The other baselines tested are described:

\begin{figure*}[t]
\begin{center}
\includegraphics[width=\linewidth]{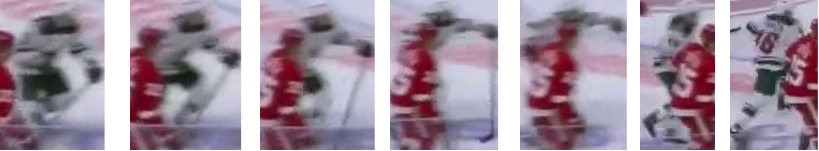}
\end{center}
  \caption{Example of a tracklet where the team is misclassified. Here, the away team player (white) is occluded by the home team player (red), which causes the team identification model to output the incorrect result. Since the original tracklet contains hundreds of frames, only a subset of tracklet frames is shown.  }
\label{fig:team_error}
\end{figure*}

\begin{enumerate}
  \item  Majority voting: after filtering tracklets based on jersey number presence, each window probability $p_i \in P$ for a  tracklet is argmaxed to obtain window predictions after which a simple majority vote is taken to obtain the final prediction. For post-processing, the majority vote is only done for those window predictions with are not the $null$ class. 
    \item  Only averaging probabilities: this is equivalent to our proposed approach without  visibility filtering and post-processing.

\end{enumerate}

The results are shown in Table \ref{table:logit_averaging}. We observe that our proposed aggregation  technique performs the best with an accuracy of $83.17\%$ and a macro weighted F1 score of $83.19\%$. Majority voting shows lower performance with accuracy of $80.59\%$ even after the visibility filtering and post-processing are applied. This is because majority voting does not take into account the overall window level probabilities to obtain the final prediction since it applies the argmax operation to each probability vector $p_i$ separately. Simple probability averaging without visibility filtering and post-processing obtains a $7.53\%$ lower accuracy demonstrating the advantage of visibility filter and post-processing step. The proposed method without the post-processing step lowers the accuracy by $2.37\%$ indicating post-processing step is of integral importance to the overall inference pipeline. The proposed inference technique without visibility filtering performs poorly when post-processing is added with an accuracy of just $50.10\%$. This is because performing post-processing on every tracklet irrespective of jersey number visibility  prevents the model to assign the $null$ class to any tracklet since the logits of the $null$ class are never taken into aggregation. Hence, tracklet filtering is an essential precursor to the post-processing step.


\subsection{Overall system}
We now evaluate the  holistic pipeline consisting of player tracking, team identification, and player identification. This evaluation is different from evaluation done the Section \ref{subsection:player_identification} since the player tracklets are now obtained from the player tracking algorithm (rather than being manually annotated). The accuracy metric is the percentage of tracklets correctly classified by the algorithm.

Table \ref{table:pipeline_table} shows the holistic pipeline. Taking advantage of player roster improves the overall accuracy for the test videos by $4.9\%$. For video number $11$, the improvement in accuracy is almost $24.44 \%$. This is because the vectors $v_a$ and $v_p$ help the model focus only on the players present in the home and away roster. There are three main sources of error: \begin{enumerate}
    \item Tracking identity switches, where the same ID is assigned to two different player tracks. These are illustrated in Fig. \ref{fig:id_switch};
    \item Misclassification of the player's team, as shown in Fig. \ref{fig:team_error}, which causes the player jersey number probabilities to get multiplied by the incorrect roster vector; and
    \item Incorrect jersey number prediction by the network.
\end{enumerate} 

\begin{table}[!t]

    \centering
    \caption{Overall player identification accuracy for 13 test videos. The mean accuracy for the video increases by $4.9 \%$ after including the player roster information }
    \footnotesize
    \setlength{\tabcolsep}{0.2cm}
    \begin{tabular}{c|c|c}\hline
  
       Video number & Without roster vectors & With roster vectors\\\hline
     
     1  & $90.6 \%$ &  $\textbf{95.34\%}$ \\
      2  & $57.1\%$ &  $\textbf{71.4\%}$ \\
       3  & $84.2\%$ &  $\textbf{85.9\%}$ \\
        4  & $74.0\%$ &  $\textbf{78.0\%}$ \\
        5   & $79.6\%$ &  $\textbf{81.4\%}$ \\
         6  & $88.0\%$ &  $88.0\%$ \\
      7  & $68.6\%$ &  $\textbf{74.6\%}$ \\
       8  & $91.6\%$ &  $\textbf{93.75\%}$ \\
        9  & $88.6\%$ &  $\textbf{90.9\%}$ \\
        10   & $86.04\%$ &  $\textbf{88.37\%}$ \\
         11  & $44.44\%$ &  $\textbf{68.88\%}$ \\
        12  & $84.84\%$ &  $84.84\%$ \\
        13   & $75.0\%$ &  $75.0\%$ \\ \hline
        Mean &  $77.9\%$  & $\textbf{82.8\%}$
   
    \end{tabular}
    \label{table:pipeline_table}
\end{table}

\section{Conclusion}
We have introduced and implemented an automated offline system for the challenging problem  of player tracking and identification in ice hockey. The system takes as input broadcast hockey video clips from the main camera view and outputs player trajectories on screen along with their teams and identities.  However, there is room for improvement. Tracking players when they leave the camera view and identifying players when their jersey number is not visible is a big challenge.
In a future work, identity switches resulting from camera panning can be reduced by tracking players directly on the ice-rink coordinates using an automatic homography registration model \cite{jiang2020optimizing}. Additionally player locations on the ice rink can be used as a feature for identifying players.


%
\IEEEpeerreviewmaketitle

%
%
%
%



%



\section*{Acknowledgment}

This work was supported by Stathletes through the Mitacs Accelerate Program and Natural Sciences
and Engineering Research Council of Canada (NSERC). We also acknowledge Compute
Canada for hardware support.

\ifCLASSOPTIONcaptionsoff
  \newpage
\fi



%



{\small
\bibliographystyle{plain}
\bibliography{egbib}
}
\end{document}